%% file: iconicity.tex
\documentclass[runningheads]{llncs}
\usepackage{graphicx}
\usepackage{amsmath,amssymb} % define this before the line numbering.
\usepackage{color}\usepackage[width=122mm,left=12mm,paperwidth=146mm,height=193mm,top=12mm,paperheight=217mm]{geometry}
\begin{document}

% \linenumbers

%%%% ADDITIONAL COMMANDS
\newcommand{\rk}[1]{{\color{blue}{#1}}}
\newcommand{\TODO}[1]{{\color{red}{\bf ToDo: #1}}}
\newcommand{\RR}{I\!\!R} %real numbers
\newcommand{\sembrack}[1]{[\![#1]\!]}
\newcommand{\myparagraph}[1]{\noindent {\bf #1.}}
\newcommand{\ie}{{\it i.e. }}
\newcommand{\eg}{{\it e.g. }}

\pagestyle{headings}
\mainmatter
\title{What makes an Image Iconic?\newline A Fine-Grained Case Study} % Replace with your title
\author{Yangmuzi Zhang$^{\dagger}$, Diane Larlus$^{\star}$, and Florent Perronnin$^{\star}$}
\titlerunning{What makes an image iconic?}
\authorrunning{Y. Zhang, D. Larlus, F. Perronnin}
\institute{$^{\dagger}$University of Maryland, College Park, USA\newline$^{\star}$Xerox
  Research Centre Europe (XRCE), Meylan, France}

\maketitle

%%%%%%%%% ABSTRACT

\begin{abstract}
\input{abstract}
%\keywords{Iconicity, Fine-Grained recognition}
\end{abstract}

%%%%%%%%% BODY TEXT
\input{intro}

\input{relatedwork}

\input{dataset}

\input{criteria}
\input{experiments}
\input{conclusion}

\bibliographystyle{splncs03}
\bibliography{iconicity}

\end{document}

%% file: abstract.tex
A natural approach to teaching a visual concept,
\eg a bird species, is to show relevant images.
However, not all relevant images represent a concept equally well.
In other words, they are not necessarily iconic.
This observation raises three questions. 
Is iconicity a subjective property? 
If not, can we predict iconicity? 
And what exactly makes an image iconic?
We provide answers to these questions through an extensive experimental 
study on a challenging fine-grained dataset of birds.
We first show that iconicity ratings are consistent across individuals,
even when they are not domain experts, thus demonstrating that iconicity is not purely subjective.
We then consider an exhaustive list of properties that are intuitively related to
iconicity and measure their correlation with these iconicity ratings.
We combine them to predict iconicity of new unseen images.  
We also propose a direct iconicity predictor that is discriminatively trained with iconicity ratings.
By combining both systems, we get an iconicity prediction that approaches human performance.

%% file: intro.tex
\section{Introduction}
\label{sec:intro}

%It has been shown through experimental studies that 
Humans often associate a concept, \eg an object, a scene, a place or a sentiment,
with a normalized visual representation, referred to as a canonical representation.
This observation motivated the introduction of the notion of a \textit{canonical} or \textit{iconic} image:
an image is said to be canonical/iconic w.r.t a given concept if it is a ``good representative'' for the said concept. 
%Several characteristics can be viewed as indications of ``representativeness''. In~\cite{Blanz1996}, for instance, an image is iconic if it is 
%(1) the best liked image of the concept, 
%(2) the picture one would see when imagining the concept,
%(3) the photo one would take of the concept or 
%(4) the image that facilitates recognition. 
%Similar to (4), we define as iconic \textit{an image that one would show, for instance to a child,
%to teach a concept}, following Berg~\cite{Berg2009}. %, in our study.
%which can be loosely defined
%as an image which is a \textit{good representative} of the concept it depicts.
% of the concept it depicts.
Rosch and Palmer~\cite{Palmer1981} showed in their seminal work that humans
agree on canonical views of objects, and that recognition is faster for these views.

Several works have considered the task of predicting image
\textit{iconicity}~\cite{Berg2009,Berg2007,Ehinger2011a}
\cite{Jing2007,Li2008,Mezuman2012,Raguram2009,Weyand2011}. % with respect to a concept.
Such iconicity predictors have many applications.
In the graphics domain, this can be used to choose the best illustration of a concept~\cite{Garg2011}.
For image search on the web, iconicity can be exploited to rerank the top-retrieved results~\cite{Jing2007,Raguram2009}. 
%For product search, this would help selecting the best shot of a product \cite{}.
In the case of a consumer photo application, it allows summarizing a collection of photographs, 
\eg a set of holiday pictures~\cite{Simon2007,Weyand2011}.
Finally, for a semi-automatic visual annotation system which involves a
human in the recognition loop~\cite{Branson2010,Wah2011iccv},
iconicity prediction enables choosing the ``best'' images to display,
\ie those that will help annotators making their decision. 
We are particularly interested in this last scenario, and more precisely in 
difficult visual recognition tasks that
require expert knowledge but are crowd-sourced to non-expert annotators.
This includes fine-grained recognition tasks which involve a large number of visually similar 
and semantically related classes (\eg species of birds and flowers, brands and makes of vehicles, etc.).
In such a case, annotators cannot solely rely on class names (\eg 'Barn Swallow') or descriptions that are very technical.
They need to be provided with appropriate -- iconic -- visual representations. 
To our knowledge, the question of how to choose such
iconic images for annotation interfaces has been largely overlooked by the computer vision community.
In this work, we raise three questions.
\textit{Is iconicity a purely subjective property?}
\textit{Can we predict an image's iconicity with respect to a concept?}
\textit{What makes an image iconic?}
We provide answers in a fine-grained classification scenario.

%

%Such iconic classifiers have many applications including choosing the best
%illustration for a concept in the graphics domain~\cite{Garg2011}, reranking the
%top-retrieved results for web image search~\cite{Jing2007,Raguram2009},
%summarizing photographs collections (\eg holiday
%pictures)~\cite{Simon2007,Weyand2011}, and providing images to facilitate the
%recognition task in a semi-automatic visual classification
%system~\cite{Branson2010} if the textual description is not enough. 

%To answer both questions, we conducted an in-depth experimental study.
%The first requirement of our study is a more precise definition of ``iconic''.
%In line with our primary target application -- visual recognition with a human in the loop --
%we follow Berg~\cite{Berg2009} and define as iconic
%\textit{an image that one would show, for instance to a child, to teach a concept}.

\begin{figure}[t!]
\centering
\includegraphics[width=1\linewidth]{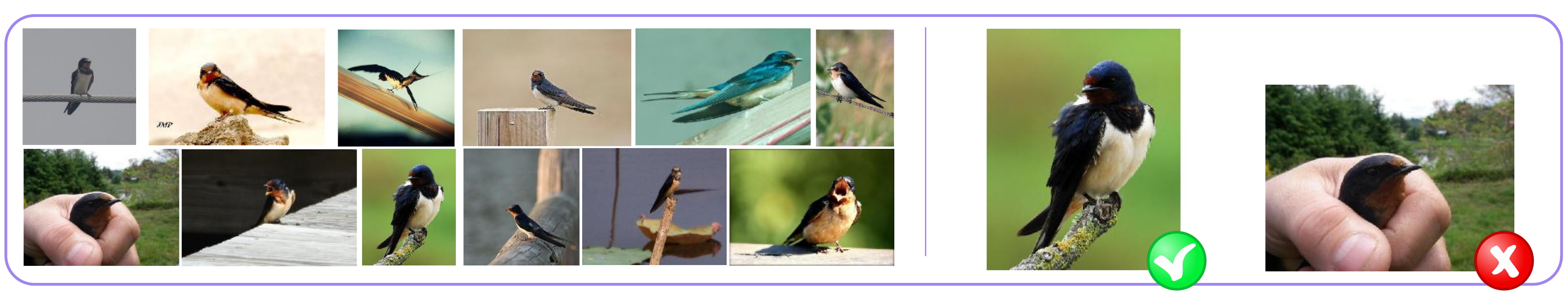}
\vspace{-0.7cm}
\caption{Given the set of images on the left, which one would you use to teach what a \textit{Barn Swallow} is?
  The picture with the green check mark is selected automatically by our method.
  On the contrary, the picture with the red cross is regarded as the least suitable.
  The rationale for these decisions is explained in Fig.~\ref{fig:qualitative}.}
\label{fig:ex}
\vspace{-0.3cm}
\end{figure}

%The first requirement to conduct such a study is to clarify the rather ambiguous 
%definition of iconicity.
%Indeed, many criteria can be used to reflect ``representativeness''.
%For instance, an image can be referred to as iconic if it is \cite{Blanz1996}
%(1) the best liked image of the concept, 
%(2) the picture one would see when imagining the concept,
%(3) the photo one would take of the concept or 
%(4) the image that facilitates recognition.
%The definition we use is closest to (4) and is consistent with a visual recognition task.
%More precisely, we follow Berg and Berg~\cite{Berg2009}
%and \textit{define as iconic an image that one would show, for instance to a child,
%to teach a concept}.

%Toward this goal, we collected a set of annotations (iconicity scores) and conducted a study

The first requirement for our study is a dataset.
Since our focus is on fine-grained tasks and we are not aware 
of any publicly-available fine-grained dataset with iconicity scores, 
we leverage the CUB dataset~\cite{Wah2011} 
which contains images of 200 bird species.
We collected iconicity scores from non-expert annotators. 
In line with our primary target application -- recognition with a human in the loop --
they were asked to choose the images that they would show to teach what a bird species looks like.
%We emphasize that {\em iconicity is a relative property} that should be measured
%for a particular category and applicative context. 
We note that the level of expertise of the annotator is likely to play a role
in rating image iconicity.
%and that
%whether an image is iconic might vary from one population of users to another, 
%\eg expert {\em vs.} non-expert.
While studying what makes an image iconic for experts would also be interesting, 
we believe there is much value in studying this problem when the users are non-experts
since the majority of annotators encountered on crowdsourcing platforms
are not.

The second requirement is to establish a large list of properties
that may play a role in deciding whether an image is iconic.
This includes the object size and position, the visibility of its parts 
or the presence of the attributes associated with the class,
the aesthetics and the memorability of the image,
the similarity of this image to the average representation of its class,
and its discriminability w.r.t. its class.
%its centrality with respect to the other images in the concept
%and its discriminability w.r.t. its class.
%For many of these properties, 
These properties are quantified by what we later refer to as {\em iconicity
indicators}. % or simply {\em indicators}.
We also describe approaches to measuring these indicators.
We measure their correlation with iconicity.
This is in contrast with the vast majority of the previous studies on the topic which 
usually focus on a single property, 
for instance the viewpoint (see section \ref{sec:rw} for a review of %the 
related work), or provide only qualitative results.

Finally, we consider iconicity predictors directly trained on generic image
descriptors using iconicity labels. 
To the best of our knowledge, such a direct approach to iconicity
prediction has never been considered.
Despite its simplicity, 
we show that it yields surprisingly competitive results and that it is complementary
to the indicator-based approach.

In summary, the contributions of this work are manyfold.
First, we enrich an existing public dataset with iconicity ratings 
%collected from a set of users 
%(section \ref{sec:dataset}) 
and show that agreement between annotator ratings is significant,
\ie that iconicity is not a purely subjective property.
%which we intend to make public.
Second, we propose an extensive list of properties likely to be relevant
to predict iconicity 
%(section \ref{sec:criteria})
and measure the correlation between these properties and the user iconicity ratings.
%(section \ref{sec:analysis}). 
To the best of our knowledge, this is the first
quantitative study of its kind.
Third, we study the combination of these different properties to 
predict the iconicity of a new image. % (section \ref{sec:predicting}). 
Fourth, we propose a direct approach to iconicity prediction %(section~\ref{sec:dip})
and show that it obtains competitive results.
Combining all our predictors we achieve a prediction accuracy that approaches the 
human performance upper-bound.
Finally, we provide qualitative results showing that the conclusions we draw
from bird images can be extrapolated to very different types of objects, such as planes and shoes.

%% file: relatedwork.tex
\section{Related Work}
\label{sec:rw}
In this section, we review the properties that have been used in
previous works to predict iconicity.
%Some papers confuse criterion and definition. 
%Most previous works %on iconic images
Mainly two properties have been considered:
the viewpoint or the ability to summarize a collection. 
Only few works considered properties beyond 
viewpoint and summarization, and combined these properties.
We also outline the limitations of previous evaluations
which were mostly qualitative. % or used an indirect evaluation protocol.

%\vspace{2mm}

\myparagraph{Iconicity and viewpoint}
Following the seminal work of~\cite{Palmer1981} that showed evidence of a
consistently preferred viewpoint, many works studied the link between iconicity
and viewpoint, and
considered different photos of the same object instance,
typically viewed under ideal conditions (\eg synthesized object with no background).
%Under such conditions, one of the main factors impacting the object appearance is the viewpoint.
Several user studies have verified the existence of iconic viewpoints
for 3D objects~\cite{Blanz1996,Bulthoff1992} as well as for scenes~\cite{Ehinger2011}.
Several works have also considered the problem of computing the best viewpoint
%B\"ulthoff and Edelman~\cite{Bulthoff1992} provided evidence that the process of recognizing a 3D
%object involves a set of 2D canonical views.
%Blanz \etal~\cite{Blanz1996} considered a visual photograpgy task and a mental imagery task
%and showed that there are clear patterns of preferred views
%and that they differ depending on the task.
%Ehinger and Oliva~\cite{Ehinger2011} considered canonical views of scenes
%and demonstrated experimentally that people have a preference for 
%views that show as much of the surrounding space as possible.
from a 3D model~\cite{Weinshall1994} or a set of 2D shapes~\cite{Denton2004}.
%which are most dissimilar to each other while being as similar as possible to the non-canonical views.
%and proposed a semidefinite program formulation to find such views.
We agree that the viewpoint plays an
important role in finding a good representative for a category,
but argue that other properties should also be taken into account
in the definition of iconicity in more realistic scenarios. 

%\vspace{2mm}

\myparagraph{Iconicity and summarization}
Many works also considered the case where the image set is 
a large collection of noisy images collected from the Internet,
for instance by querying a search engine such as Google Image Search
or a photosharing website such as Flickr for a specific concept.
In this scenario, an iconic image is an image that best summarizes the data and the 
problem of finding iconic images is generally treated as one of finding clusters 
\cite{Jing2007,Li2008,Raguram2009} or modes \cite{Mezuman2012,Weyand2011}
in the image feature space.
In most of these works, 
the results are evaluated either qualitatively through a manual inspection 
of the found iconic images \cite{Mezuman2012,Raguram2009,Weyand2011}
or simply by measuring whether the found iconic images are relevant or not
with respect to the concept \cite{Li2008}.
Yet, according to our definition, a relevant image is not necessarily iconic.
Only Jing et al. \cite{Jing2007} conducted a user study to evaluate whether the found
iconic images were preferred to random images.

%\vspace{2mm}

\myparagraph{Beyond viewpoint and summarization}
%Other criteria have been suggested to go beyond the cluster/mode criterion.
%As already mentioned, unfortunately no user study was conducted in \cite{Raguram2009}
%and it is impossibe to know what is the contribution of the aesthetic factor
%to the end result.
Berg and Forsyth~\cite{Berg2007} proposed a nearest-neighbor classifier to predict 
image iconicity and used 
figure/ground segmentation to focus on the foreground.  
Images of landmarks collected on Flickr were evaluated by 23 volunteers as
representing well the landmark or not. However, their study focuses on instances,
not categories, and does not provide any detailed analysis
as to what makes an image iconic.
Berg and Berg~\cite{Berg2009} suggested properties that could correlate
with iconicity such as the object size and position.
However, in their evaluation, the users were explicitly instructed 
to take these criteria into account which biased the results somewhat 
favorably toward these properties.
Raguram and Lazebnik~\cite{Raguram2009} proposed to leverage an aesthetic measure 
but only a qualitative evaluation of the impact of the aesthetic factor was conducted.
%of the object as well as the composition of the photograph.
Ehinger et al~\cite{Ehinger2011a} showed that images typical for scene
categories were classified correctly with high confidence.
Our study considers properties suggested in \cite{Berg2009,Berg2007,Raguram2009}
but also new ones, and reports a detailed analysis
of the correlation between each property and iconicity. We also combine 
complementary indicators and propose a full quantitative study. 
Note that direct iconicity prediction has never been considered in the past.

%\vspace{2mm}
\myparagraph{Beyond naming}
While much work in the computer vision community has been devoted to {\em naming} objects and scenes,
more and more recent works have proposed to {\em describe} them according to
their parts and attributes \cite{ferrari2007,Lampert2009,farhadi2009}, 
their aesthetic value \cite{Ke2006,Murray2012} or their memorability \cite{Isola2013}.
Our quantitative evaluation of iconicity fits in this line of research
%shares similarities with these works
as it goes beyond naming objects and scenes.
Also, we measure how these properties -- attributes, aesthetics, memorability -- 
correlate with iconicity and use them as indicators in our study.

%% file: dataset.tex
\section{Dataset}
\label{sec:dataset}

\begin{figure}[t!]
\begin{center}
  \includegraphics[width=1\linewidth]{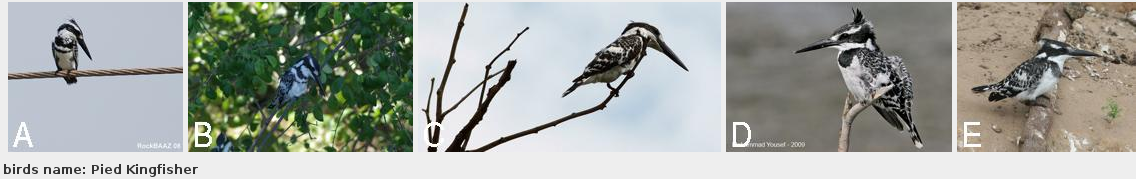}
 % \vspace{-3mm}
\vspace{-0.6cm}
  \caption{Annotators rated the iconicity of images in sets of 5.}
  \label{fig:UI}
\vspace{-0.6cm}
\end{center}
\end{figure}

We base our study on the Caltech-UCSD-Birds-200-2011 dataset (CUB for short) \cite{Wah2011},
which contains 11,788 images of 200 bird species\footnote{In this paper we use the words
  ``species'' and ``classes'' interchangeably.}.
We chose CUB for the following reasons.
First, bird species recognition is a fine-grained task, and is extremely challenging 
both for computers and non-expert annotators. 
For such problems, hybrid annotation
systems that involve a user in the recognition loop
\cite{Branson2010,Wah2011iccv} have been explored.
As explained earlier, non-expert annotators cannot solely rely on the bird names or descriptions to make decisions.
Choosing the proper iconic images is likely to make a vivid difference in such hybrid systems.
Second, the dataset contains realistic images of birds in the wild.
Although images usually have good resolution, 
by far not all of them can be considered iconic:
see Fig.~\ref{fig:ex}, \ref{fig:UI} and \ref{fig:qualitative}.
Third, this dataset comes with a rich set of annotations:
bounding boxes locate the birds precisely in images, all parts relevant
to birds (\eg beak, eyes, legs, etc.)  are indicated as visible or not, and
image-level attribute annotations describe which visual attributes can be
observed in each image.
This allows an in-depth study of a large set of iconicity indicators.
We collected iconicity ratings to enrich this
dataset\footnote{iconicity annotations available upon request} based on
the following protocol.

\myparagraph{Acquiring iconicity scores} We acquired annotations from a set of
32 non-expert annotators.  Each participant was shown 50 sets of images
corresponding to 50 bird classes and asked to rate the iconicity of each image,
where an iconic image was defined as \textit{the kind of images one would use to
  show a person what a particular bird species looks like}, in agreement with
the definition we give in the introduction.  The iconicity could be rated
according to three values: 0 for ``bad'', 1 for ``fair'', and 2 for ``good''.
For each class, 5 images of the same bird species were shown (in one row, with
the same height, see Fig.~\ref{fig:UI}).  Hence, even if the participants were
not acquainted with a particular bird species, showing multiple images
simultaneously provided them with an opportunity to get familiarized.

The dataset was initially divided by~\cite{Wah2011} into 5,994 training images,
and 5,794 test images.
For the training set, we collected annotations from 20 participants, following a
split of the data that uses some redundancy (see next paragraph), so we
obtained 4,100 iconicity annotations for the training set.
The annotated test set is composed of 2,995 images which were annotated by a
set of 12 users\footnote{15 images were annotated for each of the 200 species,
  except for 2 classes that have only respectively 11 and 14 images in the test set.}.
\textit{To avoid any bias, we made sure that participants annotating the
training and test sets were strictly different}.

We underline that {\em iconicity is not an absolute property but a relative one}:
the most iconic image of a set depends on the other images in the set.
Especially, an image might be judged more iconic with respect to a bird class
if it is compared to a set of random bird shots and less iconic
if it is compared to images which have been uploaded on Flickr (as is the case of the CUB images).
To take this fact into account, our evaluation considers relative values, \ie ranks.
This disregards global shift or scaling effects on the scores. 

\myparagraph{Consistency among annotators}
Iconicity could be seen as a subjective property.
Hence, we first conduct an experiment to measure inter-person agreement.
For that purpose, during the acquisition of the training set, 
we divided the 20 participants into 2 groups. 
All the persons in one group had 50 images (from 10 different classes) in common.
We calculate the correlation between annotations for every pair of annotators
from the same group. 

The correlation is measured using the Spearman's Rank Correlation (SRC) coefficient,
which can handle rank ties.
The $n$ scores $A_i$ and $B_i$ are converted to ranks $a_i$, $b_i$, 
and SRC is computed as:
%\vspace{-0.2cm}
\begin{eqnarray}
\rho=\frac{\sum_{i=1}^{n} (a_i -\bar{a})(b_i-\bar{b})}{\sqrt{ \sum_{i=1}^{n} (a_i
-\bar{a})^2 \sum_{i=1}^{n}(b_i-\bar{b})^2}} \in \lbrack -1, 1\rbrack
%\vspace{-0.2cm}
\label{eq:scr}
\end{eqnarray}
where $\bar{a}$ and $\bar{b}$ are the mean of the ranks $a_i$ and $b_i$
respectively. The sign of $\rho$ indicates the direction and strength of
association between the score sets $A$ and $B$, where 1 means that the two ranks
perfectly match, 0 no correlation, and -1 that they are anti-correlated.
The corresponding p-values are also calculated.
% to test the hypothesis of no
%correlation against the alternative that there is a non-zero correlation. 
If the $p$-value is small, \eg less than 0.05, it means the correlation
between annotations is significantly different from 0 with a 5\% confidence
level~\cite{Caruso1997}. 
%\TODO{DO SOMETHING}
%which indicates two annotations' correlation is strong.
For groups 1 and 2, we measured SRCs of 0.485 and 0.497 with p-values of 0.045 and 0.006 respectively.
Both p values are lower than 0.05.
Hence, despite the subjective nature of iconicity, {\em there is a strong agreement between annotators}.
This shows that {\em iconicity is not purely subjective, even when annotators are not domain experts}.
Therefore, we limited ourselves to a single iconicity score recorded per image, 
in order to cover as much of the CUB dataset as possible.
%As shown in Fig \ref{fig:qualitative}, it is possible to learn useful iconicity models
%from the collected annotations.
 %Table~\ref{tab:interConsistency} also shows the average score, and
%average standard deviation.\note{Discuss the matrix of agreement when available.}

%\begin{table}[t!]
%\begin{center}
%{\small
%\begin{tabular}{|l|c|c|}
%%\cline{2-3}
%%\multicolumn{1}{c}{} & \multicolumn{2}{|c|}{Annotator consistency}\\ 
%\cline{2-3}
%\multicolumn{1}{c|}{} & group 1 & group 2 \\
%\hline
%SRC & 0.485 & 0.497 \\
%p value & 4.50e-2 & 6.03e-3 \\
%%mean & 1.16 & 1.29 \\
%%variance & 0.3951 & 0.3496 \\
%\hline
%\end{tabular}
%}
%\caption{For 2 groups of 10 persons who annotate the same 50 images,
%  correlation between pairs of annotators.\label{tab:interConsistency}}
%%\TODO{CHANGE THIS TABLE INTO A MATRIX OF AGREEMENTS}}
%\vspace{-0.3cm}
%\end{center}
%\end{table}

\myparagraph{A few statistics}
Among the training images, 1,597 images (39\%) are rated 2, 
1,742 images (43\%) are rated 1, and 761 images (19\%) are rated 0. 
The testing set follows the same trend, with 1,161 images (39\%) rated 2, 
1,257 images (42\%) rated 1, and the rest rated 0. This shows that the full
iconicity scale was used.

%% file: criteria.tex
\section{Measuring Iconicity}
\label{sec:criteria}

We select several properties that we expect to correlate with 
image iconicity and quantify them using a variety of indicators.  Some of
these properties have been used in previous works. We also propose new ones such
as the ones based on attributes, occlusion or memorability. Each indicator
produces a score for each image.
Some of these indicators based on the available ground-truth annotations 
(\eg a bounding box) are referred to as oracles.
We leverage the rich set of annotations available with CUB for this purpose
(see previous section).
%We use as many annotations as available on this dataset to design our
%indicators.
When relevant, we also consider alternative indicators that use
predicted information (\eg from an object detector) instead of the ground-truth.
These predicted indicators can be compared to their oracle counterparts.
The properties are divided into 
i) class-independent indicators that do not need to know which class an image belongs to and
ii) class-dependent indicators that use the class label.
We finally consider the task of predicting iconicity.

\subsection{Class-independent indicators}

\myparagraph{Object size and location} 
We first look at simple statistics capturing the scene composition.  As
in~\cite{Berg2009}, we look at the object size (iconic images are expected to
present a large object) and location in the image, using the ground-truth bounding-box (BB) bird
location. We derive two indicators: \texttt{BB-size} measures the percentage of
image pixels covered by BB, and \texttt{BB-dist2center} computes the distance
between the object center (BB center) and the image center, normalized by the
length of the image diagonal.
We also study the case where the object location is unknown, and use the
state-of-the-art Deformable Part Model (DPM) object detector~\cite{Felzenszwalb2010,voc-release5}.
The DPM is trained using the BB annotations of the 200 species of birds from the
training set to build a generic bird detector.
We define two indicators computed using the DPM output instead
of the ground-truth, \texttt{DPM-size} and \texttt{DPM-dist2center}. 

%The exact same statistics can be computed using the output of the DPM
%detector instead of the ground-truth. We refer to them as \texttt{DPM-size} and
%\texttt{DPM-dist2center}. Similar statistics have been considered in~\cite{Berg2009}.

%First, we expect an iconic image to present a large object. 
%This is measured by computing the percentage of image pixels covered by BB,
%denoted as \texttt{BB-size}.
%
%The object location, which is an indicator of the quality of the composition, 
%might be an important component of iconicity as well.
%We compute the distance between the object center (BB center) and the image
%center, normalized by the image diagonal, namely \texttt{BB-dist2center}.

%If the object location is not known, one can use a detector to locate the object.
%We use the state-of-the-art Deformable Part Model (DPM) of~\cite{Felzenszwalb2010,voc-release5}.
%The DPM is trained using the BB annotations of the 200 species of birds from the training set to build% a generic
%bird detector.
%%
%The exact same statistics can be computed using the output of the DPM
%detector instead of the ground-truth. We refer to them as \texttt{DPM-size} and
%\texttt{DPM-dist2center}. Similar statistics have been considered in~\cite{Berg2009}.

\myparagraph{Occlusion}
Images in CUB are annotated with the location of 15 bird
body-parts\footnote{Parts are: both eyes, the forehead, the crown, the bill, the
  nape, the throat, the breast,
  the back, both wings, the belly, both legs and the tail.}. 
For each part, we know whether it is visible or not.
We use this information as an occlusion indicator, where the \texttt{Occlusion} score is simply the
number of visible parts.
Although view-point seems to have played a crucial role in several previous studies on
iconicity in constrained scenarios (see related work), 
it is unclear how to define such a criterion when dealing with realistic images, 
and especially with articulated objects, as is our case.
Indeed if the body of the bird faces one direction, the head can face another. 
%Hence, the notion of viewpoint is ill-defined.
Instead of building a view-point estimator that would be ill-defined in our case
and probably of low accuracy, we
use the occlusion criterion as a proxy.
% for the-view point. For instance, the fact that
%both eyes are visible gives some indication on the bird orientation.

\myparagraph{Aesthetic scores}
Although aesthetics and iconicity are not explicitly related,
we expect images of high aesthetic quality to have a higher chance to be considered iconic. 
This is because visually pleasing images are generally of high quality and well-composed~\cite{Ke2006},
and aesthetic criteria influence choosing a representative for teaching purposes (our scenario).
This intuition had already been proposed by~\cite{Raguram2009}, but only
evaluated qualitatively. We evaluate it quantitatively in our work. 
Our aesthetic predictor is based on~\cite{Murray2012}, 
which trains a classifier directly on a patch-based image representation.
This approach was shown to implicitly capture the photographic rules which are explicitly 
encoded by~\cite{Ke2006}, while providing a superior performance.
As suggested in~\cite{Murray2012}, we use Fisher Vector (FV) 
image representations~\cite{Sanchez2013} % to predict the aesthetic quality of an image
(see section~\ref{sec:exp} for more details).
To train our aesthetic indicators, we leverage the large AVA dataset~\cite{Murray2012}
which contains $\sim$200,000 images labeled with binary aesthetic labels (high
or low aesthetic quality).
We first consider a generic model, trained with the full training set. 
This model is then applied to the bird images, and the score of an image is simply the SVM score. 
This corresponds to indicator \texttt{Aesthetic-Generic}. 
A subset of images of the AVA dataset is also annotated with semantic tags.
The most relevant tag to birds is ``animal'' ($\sim$2,500 images).
Therefore, we also trained an animal-specific aesthetic model.
This indicator is referred to as \texttt{Aesthetic-Animal}.

\myparagraph{Memorability}
Memorability measures how well an image can be remembered by a person. 
We hypothesize that memorability and iconicity share common properties. 
\cite{Isola2011,Isola2013} showed that image
memorability prediction is possible with current computer vision
techniques. Consequently, we train a memorability predictor using the SUN Memorability
dataset~\cite{Isola2011}, that contains 2,222
images labeled with memorability scores.
Again, we use Fisher Vector (FV) representations~\cite{Sanchez2013}%
\footnote{To validate our feature choice, we reran the experiment in~\cite{Isola2013} and obtained a correlation similar
to the one they report (without the semantic attributes).}.
The memorability scores from the SUN dataset are used to train a linear SVM classifier. The SVM scores on the test
image constitute our \texttt{Memorability} predictor.
The dataset contains only few images with animals (less
than 10) so we did not train an animal-specific memorability model.

\subsection{Class-dependent indicators}
%\rk{do we really need more details on design choices ?}

We now consider as iconicity indicators for a given image those properties which quantify 
the relevance of the image with respect to its class.
%In this case study, we assume that the class-label of each image is known. In
%what follows we propose indicators that use this information.
%which fits the scenario we consider in our study.
%In what follows, 
We therefore assume a labeled training set of $n$ images $L=\{(x_1,y_1),(x_2,y_2),...,(x_n,y_n)\}$, 
where $x_i$ is the feature vector of image $i$ and $y_i$ is its label, 
where $y_i\in{\{1,...,K\}}$ and $K$ is the number of classes ($K=200$ for CUB).

\myparagraph{Distances to the cluster center} As mentioned in
section~\ref{sec:rw}, most previous works treated the problem of finding iconic
images as one of finding clusters or modes. Our first set of indicators follows
related work.
Since we have very few training images per class (on the order of 30), it is
unreasonable to run a clustering algorithm per class. Therefore, we
assume that each class $k$ has a single cluster whose mean is denoted $\mu_k$.  For any new
image $x'$ from the test set, we compute a similarity to the cluster center
($-||x'-\mu_k||^2$), and use this score as iconicity measure for class $k$.
Following~\cite{Li2008,Raguram2011}, we first consider the GIST
descriptor~\cite{Oliva2001} as an image feature vector.
This indicator is denoted \texttt{Cluster-GIST}.
The average GIST descriptor of a class was also used as a semantic
representative in~\cite{Raguram2009,Li2008}.
%We could have used several modes, but due to the
%small number of images per class, we decided to extract only one. 
We also consider the Fisher-Vector (FV)~\cite{Sanchez2013} representation
which results in the \texttt{Cluster-FV} indicator.
%We chose the FV because it has been shown to be a state-of-the-art representation
%in retrieval~\cite{Jegou2012} and classification~\cite{Sanchez2013} tasks.

\myparagraph{Object classifier scores} An image that receives a high score from
a classifier trained to recognize one class should represent this class well
because it is supposed to contain more discriminative features.  Consequently,
we consider using classifier scores as one of our indicators.  This was also
used to measure scene typicality of images in~\cite{Ehinger2011a}.  More
precisely, we train one linear SVM classifier per class using the labeled
training set $L$.  Then the iconicity of a new image $x'$ with respect to a
class can be measured by computing the corresponding classifier score.
Again, we consider two types of descriptors: the GIST
descriptor~\cite{Oliva2001} (indicator denoted~\texttt{SVM-GIST}) and the
Fisher Vector~\cite{Sanchez2013} (\texttt{SVM-FV}).

\myparagraph{Classifiers using attributes}
CUB contains annotations for $M$=312 attributes\footnote{The 312 attributes describe the bird
color (of the wings, the back, the forehead, etc.) and shape (of the bill, the
 tail, etc.).}, at the image level. In other words, 
each image is associated with an attribute representation $a=[a_1,a_2,...,a_M]$,
where $a_i$ takes binary values to indicate whether an attribute is present
in this image or not.
All the remaining indicators are based on the intuition that %attributes are
%related to iconicity in that 
an iconic image for a given class should best
display the attributes of that class. To the best of our knowledge, this is the
first time that this criterion has been considered and evaluated for image
iconicity.  %To validate this hypothesis, 
We considered 4 different indicators based on attributes.

%Two sets of attribute label vectors are available with the dataset. 
%The first one is an image-level attribute ground-truth that indicates whether or
%not an attribute is present in an image\footnote{The 312 attributes describe the bird
%  color (of the wings, the back, the forehead, etc.) and shape (of the bill, the
%  tail, etc.)}. 
%Each image is associated with an attribute
%representation $a=[a_1,a_2,...,a_M]$ and $M$ is the number of attributes. All
%attributes have binary values $0,1$ to indicate whether an attribute is
%present in an image or not. 

%We have image level attribute ground truth indicating whether or not an attribute is presented in an image. Each image is associated with an attribute representation $a=[a_1,a_2,...,a_M]$ and $M$ is the number of attributes. All attributes have binary values $0,1$ to indicating whether an attribute is presented in an image or not. Suppose $a^c=[a_1^c,a_2^c,...,a_M^c]$ is the binary class level attribute label.

%In another word, $n$ instances $L=\{(x_1,y_1),(x_2,y_2),...,(x_n,y_n)\}$, where $x_i$ is the attribute representation of image $i$ and $y_i\in{(1,2,...,K)}$, $K$ is the total number of classes. We train a SVM classifier. Based on this SVM classifier, we could obtain score for each image. 

%\myparagraph{KNN model}
First, we can use these image-level attribute annotations together with a
distance-based classifier. Let us assume that we have a class-level
attribute vector: $c=[c_1,c_2,...,c_N]$ (built by averaging image-level attribute
vectors).
We define an image-to-class similarity (I2C) between an image and class
$C$ as: $-||a-c||$. This similarity is used as our indicator score
and is referred to as \texttt{I2C-Att-Orac}.
%As mentioned, class-level attribute annotations are also available with CUB.
%To the best of our understanding, these class-level attribute vectors were
%obtained as the union of the relevant image-level annotations.
%The indicator that leverages these annotations is referred to as \texttt{I2C-Att-Orac-U}.
%However, we observed these annotations to be very noisy.
%
%Consequently, we also considered the average of the image-level
%attribute vectors of training images of one class as the new class-level attribute
%representation. This will be our \texttt{I2C-Att-Orac-A} indicator.
%This indicator can be understood as a ``distance to cluster center'' indicator
%but computed in the attribute feature space.

We also use the image-level attribute vectors as image representations, and we
directly train SVM classifiers on top to recognize bird
species, using the attribute vectors of the training set.
Then the trained per-class classifier can be used to predict a score for
each image. We denote this indicator \texttt{SVM-Att-Orac}. 

%\myparagraph{DAP model}
Finally, the DAP~\cite{Lampert2009} model is a standard way to predict categories
based on attribute-level information. Given image $x$, we first obtain 
the attribute predictions $p(a_m|x)$ by training $M$
independent attribute classifiers. The score of image $x$ is then given by
%\begin{eqnarray}
%\label{eqn:eqnDAP}
$p(a|x)=\prod_{m=1}^Mp(a_m|x)$.
%\end{eqnarray} 
We considered an oracle scenario and a prediction one.
For the oracle scenario, the probabilities are 0 or 1, based
on the image level annotation. 
We use $\epsilon$=$10^{-5}$ for probability $0$ and 1-$\epsilon$ for probability $1$ to avoid the overall $p(a|x)=0$. 
This indicator is referred to as \texttt{DAP-Orac}.
%As was the case before, we consider two indicators depending on
%whether we use the union or the average of the per-image
%attributes to compute the class-level attribute representations (\texttt{DAP-Orac-U} and \texttt{DAP-Orac-A} respectively).
%
For the prediction scenario, we assume that test images are not annotated with
image-level attribute vectors, and we predict attribute
probabilities using attribute classifiers trained on training images (\texttt{DAP-Pred}). 
%These classifiers are trained using
%the per-class image attributes (A) as it leads to better results in previous indicators. 
%
%All DAP models are learnt once again on FV representations.
All DAP models are learnt on FV representations.

\subsection{Iconicity Prediction}
\label{sec:dip}

Given the previous indicators, we can now predict iconicity.
The simplest approach consists in linearly combining the indicators, 
either by giving them equal weights, or by learning a vector of weights $w$ using iconicity labels.
In the latter case, we concatenate the indicator values,
whiten the resulting feature vector and learn a linear SVM.
We experimented with two learning frameworks: a binary SVM classifier and a ranking SVM~\cite{joachims2002}.
A disadvantage of the binary SVM formulation is that it requires setting an arbitrary threshold
that will split the training set into iconic {\em vs.} non-iconic images.
In our experiments, we set this threshold to 1.5.
The ranking SVM formulation deals directly with ranked training pairs, which
fits better the relative nature of iconicity. More formally, given training pairs of images $(x^+,x^-)$ such that
$x^+$ is ranked higher than $x^-$, we minimize the regularized ranking loss:
\begin{equation}
\sum_{(x^+,x^-)} \max\{0,1-w'(x^+-x^-)\} + \frac{\lambda}{2} ||w||^2 .
\end{equation}
Note that, in our implementation, the images $x^+$ and $x^-$ 
of a given pair $(x^+,x^-)$ are from the same batch of 5 images annotated by the same user.
This is to avoid confounding factors during the learning process 
(\eg the fact that a user might be more inclined to rate images as iconic than other users).

As an alternative to the indicator-based approach to iconicity prediction,
we also consider the approach which consists \textit{in predicting the iconicity
directly from image representations such as the FV}.
We refer to this approach as Direct Iconicity Prediction (DIP).
As is the case for the indicator-based predictor, we use a linear SVM on the FV features and its parameters 
can be learned using either a binary classification objective function or a ranking objective function.
To our knowledge, this is the first attempt to such a direct approach to
iconicity prediction.
These predictors are denoted \texttt{DIP-bin} and \texttt{DIP-rank}

%% file: experiments.tex
\section{Experiments}
\label{sec:exp}
%This experimental study considers all indicators described in the previous
%section and the direct iconicity predictor. 
After discussing implementation details (section~\ref{sec:det}), 
we first evaluate the correlation of each indicator with the iconicity ratings 
and with each other (section~\ref{sec:analysis}).
In the second set of experiments, we evaluate how well we predict iconicity
by combining different indicators and by direct prediction (section~\ref{sec:predicting}).

In all experiments we use the standard train/test split of CUB.
All supervised learning (\eg training DPMs, 
computing class means, training SVM classifiers)
is performed on the train split and all the 
results are measured on the portion of the test split 
which was rated by the users.
For validation, we split the train set into two halves:
we train on the first half, validate on the second half and retrain
the models on the full training set with the optimal validation parameters.

\subsection{Implementation details and evaluation}
\label{sec:det}

\myparagraph{Implementation details}
We use GIST and Fisher-Vector (FV) descriptors in some of the listed indicators. 
For the GIST, we used the color implementation of~\cite{Oliva2001} (960 dimensions).
We compute FV representations on top of SIFT descriptors~\cite{Lowe2004} 
and color descriptors~\cite{Clinchant2007} for each image. 
The number of Gaussians is 1,024.
We use a spatial pyramid~\cite{lazebnik2006beyond} with 8 regions:
the full image, 3 horizontal stripes and the 4 quadrants.
%\note{Add more details: PCA, nb gaussians, spatial pyramid -> final dimension,
%  and shorten what follows.}
There is one FV pipeline for each low-level descriptor, 
and both pipelines are combined with late fusion (score averaging).
%Aesthetic and memorability models are also learnt by a late fusion of SIFT- and color-FV representations.

\myparagraph{Measures}
We consider two different evaluation measures. %, that both consider relative
%values (=rank) and not absolute ones.
%
First, in accordance with the binary classification setting, 
we evaluate the quality of indicators using the average precision (AP).
We define as positive (= iconic) those images whose ground-truth label is above a threshold $\alpha$=1.5 and the
remaining are labeled as negative. 
Second, in accordance with the ranking setting, 
we look at the correlation between indicators and iconicity.
Again, it is measured using Spearman's Rank Correlation (SRC)
coefficient (see eq.~(\ref{eq:scr})).
SRC is computed between the indicator scores and
the ground-truth iconicity scores. The corresponding p-values
are also reported. A $p$-value smaller than 0.05 means
statistical correlation with a 95\% confidence level.

\subsection{Analysis of iconic images}
\label{sec:analysis}
We first evaluate the correlation of each indicator with the iconicity ratings 
and with each other.
Results are presented in Table~\ref{tab:strat}.  At
first glance, all the methods exhibit a correlation with iconicity, except
\texttt{SVM-GIST} (p-value $>$ 0.05). 
Although we can observe small differences between the ranking of the methods between
the SRC and AP measures, both of them lead to similar conclusions. 
We now provide a detailed discussion.

% include the table of results
\input{strategies_test.tex}

\myparagraph{Scene layout}
The strongest correlation is observed for the \texttt{BB-size} indicator. This
shows that the object size plays a crucial role for image iconicity. 
%To better understand this, 
We also inspected the distribution of the scores as a
function of the BB size in our training set, and observed that the scores
increase as the ratio between the BB area and full image increases up to around 0.85
and then decreases slightly. This may be because
objects that are too big become less iconic as a certain amount of context is necessary.
Also, the location predicted by the DPM detector allows producing a score that
is almost as correlated with iconicity as the ground-truth BB, showing that a
detector could successfully replace annotations in our scenario.
The distance between the object center and the center of the image exhibits a lower correlation (see 
\texttt{BB-dist2center} and \texttt{DPM-dist2center}).

%Considering the major role of object size in iconicity, we could
Considering the major role of object size, we could
have applied indicators only to the object location (\ie cropped
images).  We discarded this option as i) this is not consistent with the
%annotation acquisition process for which users were shown the full images, and
annotation acquisition (users were shown the full images), and
ii) this applies a transformation to the images, which is different from
evaluating the iconicity of an image as it is.

\myparagraph{Occlusion}
The correlation with the occlusion indicator is also high, confirming our intuition
that the number of visible parts is related to iconicity. As expected,
images of occluded objects are not iconic.
We also trained a classifier on top of the binary part-visibility vectors
to learn the importance of the parts, but could not improve the correlation.
%\note{Do we want to discuss about the occlusion-entropy that obtains: 0.0916 \&
%  0.4719 \& 7.50 ?}

\myparagraph{Aesthetic evaluation}
Both our aesthetic scores exhibit significant correlation with iconicity measures,
proving the importance of this criterion.
Even though it has been trained with much fewer 
($\sim$ two orders of magnitude fewer) images, 
the aesthetic model trained specifically with animal images 
performs better than the generic one (+0.046
correlation and +1.31 AP).

\myparagraph{Memorability}
The correlation with the memorability score is lower than with the aesthetic
one, but a correlation is still observed. This implies that %although memorability
%and iconicity are two different image characteristics, they 
memorability and iconicity share common properties.
%(see Tab \ref{tab:cpMethod}).

\myparagraph{Clustering}
Indicators computing a similarity between images and cluster centers of their class show a
significant but small correlation with iconicity, which is surprising
considering that it is the most commonly used indicator in the literature. 
The FV-based indicator  (\texttt{Cluster-FV}) performs better than the GIST one
(\texttt{Cluster-GIST}). This can be explained by the fact that the FV is a 
richer descriptor than the GIST. % that assumes a fixed geometry.

%Please note that we have extracted the corresponding ``transductive'' numbers on
%the training set, and have obtained similar results. By transductive we mean that
%images that we are interested in ranking using the score of the clustering
%indicators are also the one that are used to compute the cluster center.
%This shows that we can compute the center of the cluster based on the set of
%images whose iconicity we are interested in, but also on a different set of
%images of the same class (which would allow for offline computation).

\myparagraph{Object classifier scores}
We observe a higher correlation using the \texttt{SVM-FV} indicator. Based on the
FV representation, a discriminative classifier trained to recognize a class can
pretty well capture and predict iconic properties of that class, as already
shown by~\cite{Ehinger2011a} for different features.
Note that this classifier is not perfect (our
setting yields a top-1 classification accuracy of 32.8\% on the corresponding 200-species
classification problem) but is good enough to capture interesting properties.
On the other hand, a SVM classifier trained with GIST descriptors produces a
very low correlation. This is not surprising as the corresponding 200-class
classification results drop to 4.7\% top-1 accuracy.

%As a sanity check, we also considered GIST
%features computed on images cropped using the BB information. Aligning images
%makes more sense for this rigid descriptor, and the accuracy of the
%classification problem is brought to 11.10\% mAP.
%For this scenario (\texttt{SVM-Gist (crop)}), the obtained correlation is 0.11
%(with p-value 1.53e-09) and the AP is 45.39\%. 

\myparagraph{Attributes}
We observe that \texttt{SVM-Att-Orac} produces lower correlation than \texttt{SVM-FV}. We see
two possible explanations. First, it could be better to let the
classifier decide what a good representation of the class is from the image features  ({\em e.g.} the FVs) 
themselves than using manually chosen attributes as a proxy. Second, 
attribute annotations could be noisy as 
each image was annotated by a single AMT worker~\cite{Wah2011}.
The indicators \texttt{I2C-Att-Orac} and \texttt{DAP-Orac} perform well, but the
best attribute indicator is \texttt{SVM-Att-Orac}.

\myparagraph{Correlation between indicators}
We also measure the correlation between the different indicators.
SRC scores for a subset of the most promising indicators are
presented in Table~\ref{tab:cpMethod} (as color values).
First, we observe the high correlations obtained by all methods using the class information.  
The two most correlated methods are the \texttt{SVM-FV} and the \texttt{Cluster-FV} (this pair
obtains the very high value of 0.706), and the second highest correlation is the
attribute method together with the \texttt{SVM-FV}. 
%This high correlation is not surprising given that all these indicators rely
%on the same descriptors: FVs.
The aesthetic and the memorability indicators have very low correlation with the
other methods. In the next section, we will show that these class-independent indicators
nicely complement the class-based ones.

\subsection{Prediction of image iconicity}
\label{sec:predicting}

From the previous study, we have identified indicators that
best capture information related to image iconicity. We now show that they
are complementary.

%Based on the previous study, we have identified a list of indicators that
%capture information related to image iconicity. We now show that they
%constitute a complementary set.

\begin{table}[t!]
\begin{center}
%\begin{tabular}{ccc}
\parbox{0.40\linewidth}
{
%\begin{figure}
  \caption{Spearman's Rank Correlation (SRC) between each pair of
  selected indicators. 
%is displayed as color values.
\label{tab:cpMethod}}
  \includegraphics[width=\linewidth]{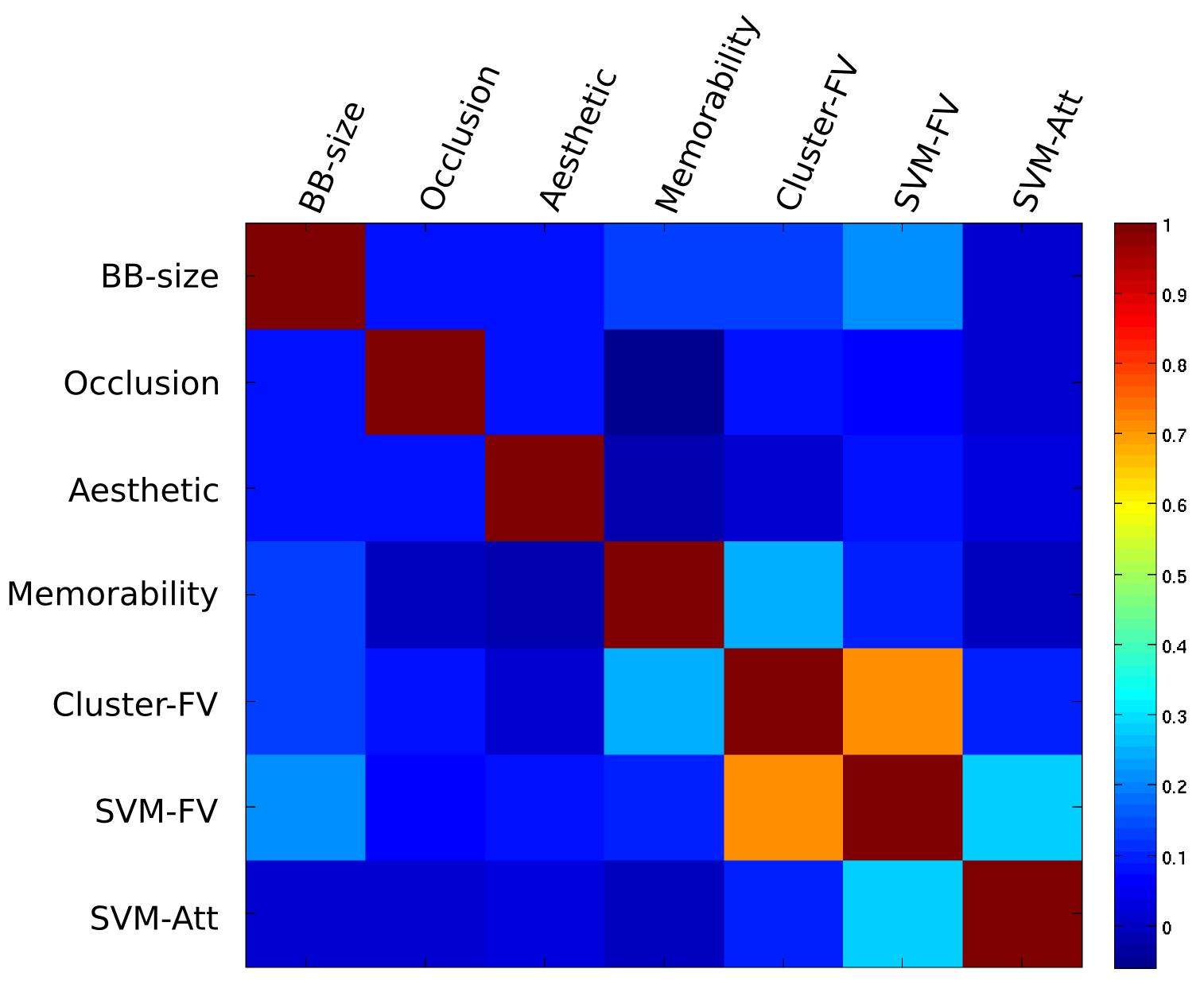}
%\end{figure}%
}
\hspace{0.3cm}
\parbox{0.55\linewidth}
{
\centering
\input{prediction_test} 
} 
%\end{tabular}
\end{center}
\vspace{-0.5cm}
\end{table}

\myparagraph{Combination of the most relevant indicators}
In a first scenario, we assume that all annotations are available, and we
combine the oracle versions of our indicators. We selected \texttt{BB-size},
\texttt{Aesthetic-Animal}, \texttt{Occlusion}, \texttt{Memorability}, \texttt{Cluster-FV},
\texttt{SVM-FV}, and \texttt{SVM-Att-Orac} due to their good correlation with
iconicity.
We also consider a second scenario, where annotations are replaced by
predictions, except for the class label that is always considered as known
(consistently with our scenario).
We combine \texttt{DPM-size}, \texttt{Aesthetic-Animal}, \texttt{Cluster-FV},
\texttt{Memorability}, \texttt{SVM-FV}, and \texttt{DAP-Pred}. There is no
occlusion criteria in this second scenario, as it seems
unrealistic to train part detectors of a high enough quality to estimate the
level of occlusion of the object.
To make them comparable, the scores of each indicator are whitened
(average 0, and std 1). 
We considered two combination methods. In the first one, we average
all scores. In the second one, indicator scores obtained for our training images
are used to train an SVM classifier that weights the different
indicators. 
Both the binary and ranking SVMs are considered. 
All results are reported in Table~\ref{tab:pred}.

\begin{figure*}[h!]
\centering
\includegraphics[width=0.92\linewidth]{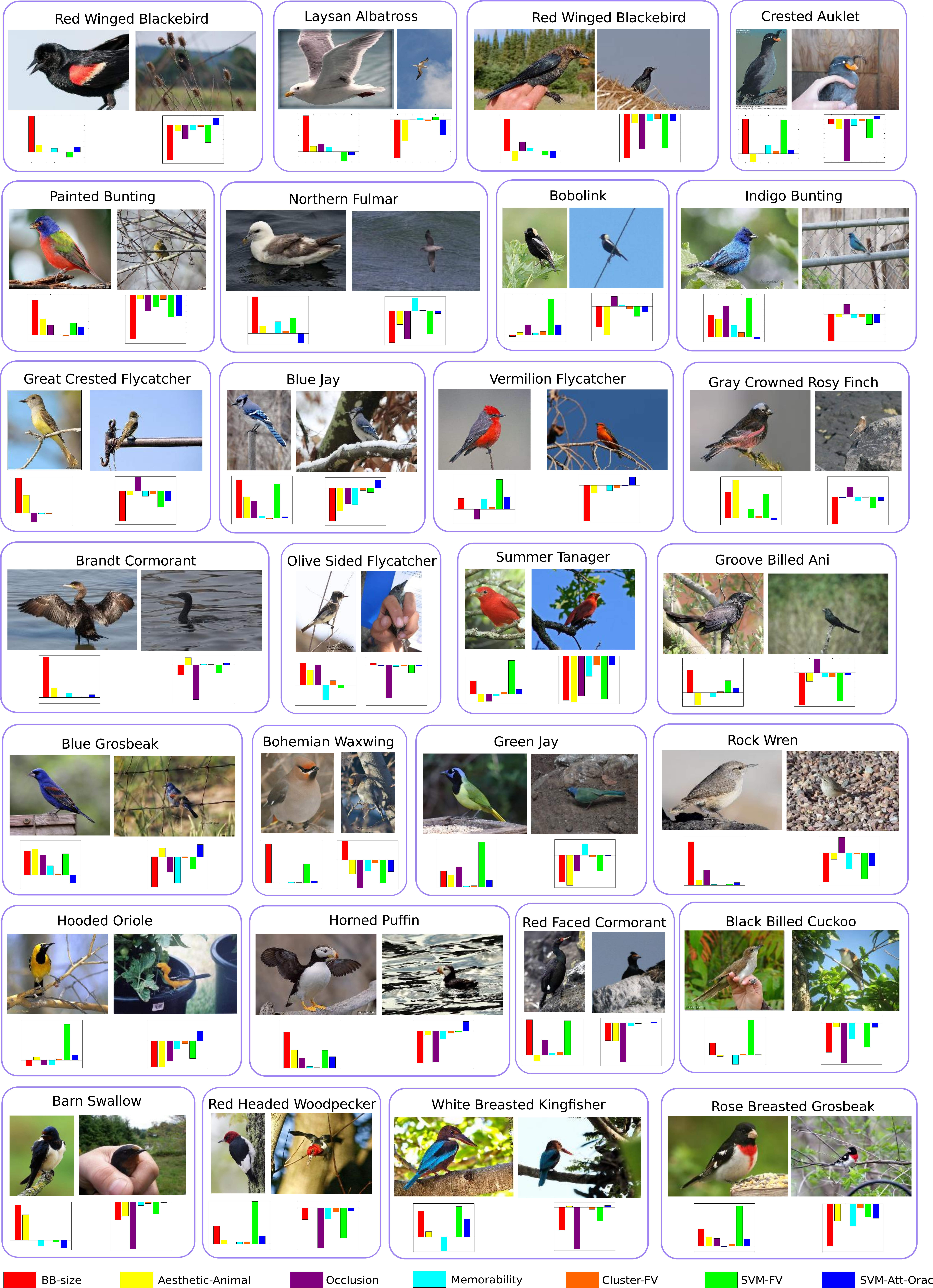}
\caption{For a set of representative bird species, images with the highest and lowest predicted iconicity
  based on our learnt combination of oracle indicators are shown
  respectively as the left and right images of each box.
  Below images, histograms display the contribution of each indicator in
  the decision process.
  For instance, for the ``Brandt Cormorant'', the \texttt{BB-size} plays a crucial role in deciding 
  the choice of the most iconic image while for the ``Summer Tanager'' almost all indicators play
  a role in the choice of the least iconic image (best viewed in color).}
\label{fig:qualitative}
\end{figure*}

First we observe that the average oracle indicator performs already
significantly better than any indicator taken independently.
This result shows that a class-independent combination works
which means that we could predict iconicity for a new bird class.
When learning a classifier, up to 2.9\% of AP (0.045 SRC) can be gained. 
Second, we see that the predicted indicators (average and learnt) 
yield results that are quite close to the oracle ones, 
demonstrating the applicability of the combination without
the optimistic oracle assumption.
Qualitative results are presented in Fig.~\ref{fig:qualitative}. 
%\note{only one or two sentences.} 
% For the Barn Swallows, the least iconic image receives a very low score on indicator \texttt{Occlusion}. 
%The bird is held in hand and most body parts are occluded. 
%The best image obtains high score from \texttt{BB-size} and
%\texttt{Aesthetic-Animal}. 
%This can be easily understood as the bird occupies a large area, and the image
%is visually pleasing. 

\myparagraph{Direct Iconicity Prediction}
The direct approach yields a high accuracy, comparable to our oracle
indicators, showing that iconicity can be directly learnt from
data. Yet, this indicator does not allow understanding what makes an image
iconic. 
%% OLD RESULTS WITHOUT MEMORABILITY
%When combining the discriminative approach and our best combination (Oracle one) we
%obtain: a correlation of 0.43 (with p-value=2.49e-137) and an AP of 63.9.
%
%If we combine the discriminative baseline with the best prediction-based
%combination, we still obtain 0.406 (with p-value=2.35e-119) correlation and
%63.0 AP which again shows that we could predict such iconicity score even on a
%scenario where less annotations are available.
When combining the best discriminative approach (\texttt{DIP-rank}) and the best
learnt oracle (\texttt{SVM-rank}), we obtain a correlation of 0.459 %(with p-value=1.e-155) 
and an AP of 64.7.
If we combine \texttt{DIP-rank} with the best prediction-based
combination, we still obtain a 0.420 %(with p-value=1.6e-128) 
correlation and
63.7 AP which again shows that we could predict such iconicity score even on a
scenario where fewer annotations are available. 
%The full result is shown in Table~\ref{tab:fvCombo}. 
The best combination we obtain, 0.459 is quite close to 
 0.485 and 0.497, that correspond to the correlation between groups of
 annotators for the same images (discussed in section~\ref{sec:dataset}).
This inter-person correlation gives an upper-bound on the maximum
prediction accuracy we can achieve, and we observe that our best combination
almost obtains this.

%\myparagraph{Google Baseline}
%When searching for visual classes on Google, the class is illustrated on the result
%page, and the first displayed image is often from Wikipedia. We assumed this
%image to be representative, and close to what an expert would have chosen.
%Consequently, we compute the similarity (as before, based on the late-fusion of SIFT and
%color FVs) between images from our test set, and this first displayed
%image\footnote{On some rare cases, the first image was not relevant (it was for
%instance a map) and we used the second image.} to illustrate that class. We
%obtain the \texttt{Google \#1} baseline.
%
%The Google baseline obtains good results (comparable to some of our individual
%indicators) but is much lower than the considered combinations. 

%\myparagraph{Predicting the most representative image}
%In Table~\ref{tab:accBest}, we predict the most iconic image using different methods.

%\input{tables/mostrep.tex}

\begin{figure*}[h!]
\centering
\includegraphics[width=0.8\linewidth]{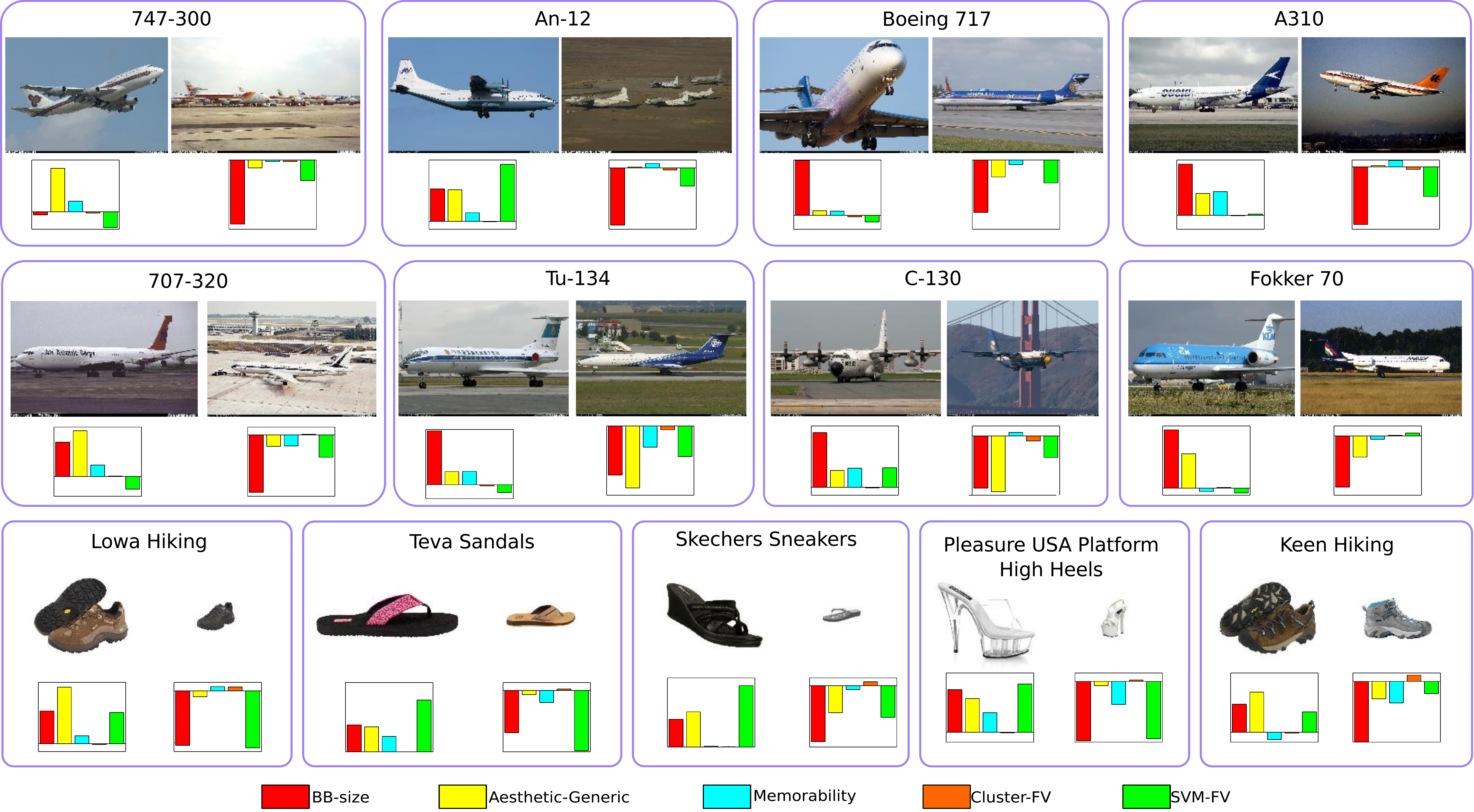}
\caption{For a set of representative classes extracted from two subsets of the FGcomp'13, namely
  shoes and planes, and for the available indicators, this figure shows
  images with the highest and lowest predicted iconicity
  based on our learnt combination of oracle indicators, 
  respectively as the left and right images of each box.
  Below images, histograms display the contribution of each indicator in
  the decision process (best viewed in color).}
\label{fig:qualitative2}
\end{figure*}

\myparagraph{Beyond birds}
We show qualitatively that our findings generalize beyond birds.
% and that the system we learned on bird images can be reasonbaly extrapolated to completely different objects.
For this purpose, we used images of the Fine-Grained Competition (FGcomp'13) \cite{fgcomp13}
and applied directly the combination of iconicity predictors learned on the bird images
to two completely different objects: planes and shoes.
Note that on these datasets, we did not have access to parts and attributes and therefore
we could not use the \texttt{Occlusion} and \texttt{SVM-Att-Orac} predictors.
The results are provided in Fig.~\ref{fig:qualitative2} and show that the chosen most/least iconic images are plausible,
although the system was trained on very different objects (birds).

%% file: strategies_test.tex
\begin{table}[t!]
\caption{Comparison of the proposed indicators on the test images. 
Spearman rank correlation (SRC), corresponding p-value, and average precision (AP) are reported.}
\vspace{-0.2cm}
\begin{center}
{\small
\begin{tabular}{|l|c|c|c|}
\hline
\multicolumn{4}{|c|}{Class-independent Indicators}\\
\hline
\hline
Strategy & SRC & p-value & AP \\
\hline
\hline
BB-size & 0.304 & 3.01e-65 & 51.7 \\
DPM-size & 0.280 & 6.84e-55 & 51.6 \\
BB-dist2center & 0.097 & 9.53e-08 & 43.4 \\
DPM-dist2center & 0.069 & 1.67e-04 & 41.3 \\
\hline
Occlusion & 0.163 & 3.17e-19 & 45.2 \\
\hline
Aesthetic-Generic & 0.139 & 2.29e-14 & 46.1 \\
Aesthetic-Animal  & 0.185 & 1.73e-24 & 47.4 \\
\hline
Memorability & 0.112 & 9.16e-10 & 43.7 \\
\hline
\end{tabular}
\hspace{0.2cm}
\begin{tabular}{|l|c|c|c|}
\hline
\multicolumn{4}{|c|}{Class-dependent Indicators}\\
\hline
\hline
Strategy & SRC & p-value & AP \\
\hline
\hline
Cluster-GIST &  0.048 & 9.20e-03 & 41.7 \\
Cluster-FV & 0.111 & 1.18e-09 & 43.3 \\
\hline
SVM-GIST  & 0.027 & 1.44e-01 & 40.9 \\
SVM-FV  & 0.233 & 2.85e-38 & 49.1 \\
\hline
SVM-Att-Orac & 0.150 & 1.90e-16 & 48.1 \\
%I2C-Att-Orac-U & 0.078 & 1.92e-13 & 42.8 \\
%I2C-Att-Orac-A & 0.126 & 5.52e-12 & 44.1 \\
%DAP-Orac-U & 0.058 & 1.50e-03 & 42.2 \\
%DAP-Orac-A & 0.113 & 5.06e-10 & 44.2 \\
I2C-Att-Orac & 0.126 & 5.52e-12 & 44.1 \\
DAP-Orac & 0.113 & 5.06e-10 & 44.2 \\
DAP-Pred & 0.063 & 5.95e-04 & 42.7 \\
\hline
\end{tabular}
}
\end{center}
\label{tab:strat}
\vspace{-0.7cm}
\end{table}

%% \begin{table}
%% \caption{Comparison of the proposed indicators on test dataset. 
%% Spearman rank correlation (SRC), corresponding p-value, and average precision (AP) are reported.}
%% \begin{center}
%% {\small
%% \begin{tabular}{|l|c|c|c|}
%% \hline
%% Strategy & SRC & p-value & AP \\
%% \hline
%% \multicolumn{4}{|c|}{Class-independent Indicators}\\
%% \hline
%% BB-size & 0.304 & 3.01e-65 & 51.7 \\
%% DPM-size & 0.280 & 6.84e-55 & 51.6 \\
%% BB-dist2center & 0.097 & 9.53e-08 & 43.4 \\
%% DPM-dist2center & 0.069 & 1.67e-04 & 41.3 \\
%% \hline
%% Aesthetic-Generic & 0.139 & 2.29e-14 & 46.1 \\
%% Aesthetic-Animal  & 0.185 & 1.73e-24 & 47.4 \\
%% \hline
%% Occlusion & 0.163 & 3.17e-19 & 45.2 \\
%% \hline
%% Memorability & 0.112 & 9.16e-10 & 43.7 \\
%% \hline
%% \multicolumn{4}{|c|}{Class-dependent Indicators}\\
%% \hline
%% Cluster-GIST &  0.048 & 9.20e-03 & 41.7 \\
%% Cluster-FV & 0.111 & 1.18e-09 & 43.3 \\
%% \hline
%% SVM-GIST  & 0.027 & 1.44e-01 & 40.9 \\
%% SVM-FV  & 0.233 & 2.85e-38 & 49.1 \\
%% \hline
%% SVM-Att-Orac & 0.150 & 1.90e-16 & 48.1 \\
%% I2C-Att-Orac-U & 0.078 & 1.92e-13 & 42.8 \\
%% I2C-Att-Orac-A & 0.126 & 5.52e-12 & 44.1 \\
%% DAP-Orac-U & 0.058 & 1.50e-03 & 42.2 \\
%% DAP-Orac-A & 0.113 & 5.06e-10 & 44.2 \\
%% DAP-Pred & 0.063 & 5.95e-04 & 42.7 \\
%% \hline
%% \end{tabular}
%% \vspace{-0.5cm}
%% }
%% \end{center}
%% \label{tab:strat}
%% \end{table}

%% file: prediction_test.tex
%\begin{table}[t!]
\caption{Combinations of indicators to predict image iconicity. The
SRC, corresponding p-value,  and average precision (AP) are reported.}
%{\small
%\begin{center}
\begin{tabular}{|l|c|c|c|}
\hline
Method               &  SRC   & p-value     & AP \\
\hline
\hline
\multicolumn{4}{|c|}{Combining Oracle Indicators}\\
%\hline
%Average of Orac (no mem)    & 0.367 & 2.64e-96  & 57.4 \\
%Learnt on Orac (no mem)    & 0.385 & 1.42e-106 & 59.4 \\
%Average on Pred (no mem)  & 0.302 & 2.50e-83 & 53.1 \\
%Learnt on Pred  (no mem) & 0.343 & 3.18e-83 & 57.9 \\
\hline
Average   & 0.370 & 5.02e-98  & 58.0 \\
SVM-bin   & 0.401 & 3.52e-116 & 60.9 \\
SVM-rank  & 0.415 & 4.94e-125 & 60.4 \\
\hline
\multicolumn{4}{|c|}{Combining Predicted Indicators}\\
\hline
Average   & 0.303 & 1.55e-64 & 53.2 \\
SVM-bin   & 0.353 & 2.23e-88 & 59.3 \\
SVM-rank  & 0.350 & 2.39e-86 & 56.4 \\
\hline
\multicolumn{4}{|c|}{Direct Iconicity Predictors}\\
\hline
DIP-bin & 0.372 & 7.34e-88 & 60.2 \\
DIP-rank  & 0.375 & 1.08e-100 & 60.3 \\
\hline
%\multicolumn{4}{|c|}{Baseline}\\
%\hline
%Google \#1    & 0.158 & 2.96e-18 & 45.9 \\
%\hline
\end{tabular}
%\end{center}
%\vspace{-0.6cm}
%}
\label{tab:pred}
%\end{table}

%% file: conclusion.tex
\section{Conclusion}
\label{sec:ccl}
The goal of this study was to verify that iconicity is an image property
rated consistently across multiple annotators 
and to understand what makes an image iconic.
%\ie a good representative of the class it depicts.
%with respect to a given class.
%where iconic is defined as a good representative of the concept if depicts.
Toward this goal, we conducted an extensive study which involved collecting
iconicity annotations from a set of users, and proposing a large set of possible
properties that can predict iconicity. These properties cover all the ones
considered in previous work, plus new ones we proposed. Some are
class-dependent, and some are totally generic. 
Our study showed that in the fine-grained context, these properties used with
the right descriptor are all relevant to iconicity, and are complementary with
each other. We also proposed to directly predict iconicity with a classifier
discriminatively trained on iconicity rating, and showed the good performance of
this simple yet novel scenario.
We expect all these findings to be useful in many computer vision applications of practical value.

%% file: iconicity.bbl
\begin{thebibliography}{10}
\providecommand{\url}[1]{\texttt{#1}}
\providecommand{\urlprefix}{URL }

\bibitem{fgcomp13}
{FGComp'13}. http://sites.google.com/site/fgcomp2013/

\bibitem{Berg2009}
Berg, T., Berg, A.: Finding iconic images. In: IVW at CVPR (2009)

\bibitem{Berg2007}
Berg, T., Forsyth, D.A.: Automatic ranking of iconic images. Tech. rep., U.C.
  Berkeley (2007)

\bibitem{Blanz1996}
Blanz, V., Tarr, M., B\"ulthoff, H., Vetter, T.: What object attributes
  determine canonical views? Tech. rep., MPI (1996)

\bibitem{Branson2010}
Branson, S., Wah, C., Babenko, B., Schroff, F., Welinder, P., Perona, P.,
  Belongie, S.: {Visual Recognition with Humans in the Loop}. In: ECCV (2010)

\bibitem{Bulthoff1992}
B\"ulthoff, H., Edelman, S.: Psychophysical support for a two-dimensional view
  interpolation theory of object recognition. PNAS  (1992)

\bibitem{Caruso1997}
Caruso, J., Norman, C.: Empirical size, coverage, and power of confidence
  intervals for spearman's rho. Educational and Psychological Measurement
  (1997)

\bibitem{Clinchant2007}
Clinchant, S., Csurka, G., Perronnin, F., Renders, J.M.: {XRCE's participation
  to ImagEval}. In: ImageEval Workshop at CVIR (2007)

\bibitem{Denton2004}
Denton, T., Demirci, F., Abrahamson, J., Shojoufandeh, A.: Delecting canonical
  views for view-based {3-D} object recognition. In: ICPR (2004)

\bibitem{Ehinger2011}
Ehinger, K., Oliva, A.: Canonical views of scenes depend on the shape of the
  space. Cognitive Science Society  (2011)

\bibitem{Ehinger2011a}
Ehinger, K.A., Xiao, J., Torralba, A., Oliva, A.: Estimating scene typicality
  from human ratings and image features. Cognitive Science Society  (2011)

\bibitem{farhadi2009}
Farhadi, A., Endres, I., Hoiem, D., Forsyth, D.: Describing objects by their
  attributes. In: CVPR (2009)

\bibitem{Felzenszwalb2010}
Felzenszwalb, P.F., Girshick, R.B., McAllester, D., Ramanan, D.: Object
  detection with discriminatively trained part based models. TPAMI  (2010)

\bibitem{ferrari2007}
Ferrari, V., Zisserman, A.: Learning visual attributes. In: NIPS (2007)

\bibitem{Garg2011}
Garg, S., Berg, T., Mueller, K.: Iconizer: A framework to identify and create
  effective representations for visual information encoding. In: Smart Graphics
  (2011)

\bibitem{voc-release5}
Girshick, R., Felzenszwalb, P., McAllester, D.: Discriminatively trained
  deformable part models, release 5.
  http://people.cs.uchicago.edu/~rbg/latent-release5/

\bibitem{Isola2011}
Isola, P., Xiao, J., Torralba, A., Oliva, A.: What makes an image memorable?
  In: CVPR (2011)

\bibitem{Isola2013}
Isola, P., Xiao, J., Parikh, D., Torralba, A., Oliva, A.: What makes a
  photograph memorable? TPAMI  (2013)

\bibitem{Jing2007}
Jing, Y., Baluja, S., Rowley, H.: Canonical image selection from the web. In:
  CIVR (2007)

\bibitem{joachims2002}
Joachims, T.: Optimizing search engines using clickthrough data. In: SIGKDD
  (2002)

\bibitem{Ke2006}
Ke, Y., Tang, X., Jing, F.: The design of high-level features for photo quality
  assessment. In: CVPR (2006)

\bibitem{Lampert2009}
Lampert, C.H., Nickisch, H., Harmeling, S.: Learning to detect unseen object
  classes by betweenclass attribute transfer. In: CVPR (2009)

\bibitem{lazebnik2006beyond}
Lazebnik, S., Schmid, C., Ponce, J.: Beyond bags of features: Spatial pyramid
  matching for recognizing natural scene categories. In: CVPR (2006)

\bibitem{Li2008}
Li, X., Wu, C., Zach, C., Lazebnik, S., Frahm, J.M.: Modeling and recognition
  of landmark image collections using iconic scene graphs. In: ECCV (2008)

\bibitem{Lowe2004}
Lowe, D.: Distinctive image features from scale-invariant keypoints. IJCV
  (2004)

\bibitem{Mezuman2012}
Mezuman, E., Weiss, Y.: Learning about canonical views from internet image
  collections. In: NIPS (2012)

\bibitem{Murray2012}
Murray, N., Marchesotti, L., Perronnin, F.: Ava: A large-scale database for
  aesthetic visual analysis. In: CVPR (2012)

\bibitem{Oliva2001}
Oliva, A., Torralba, A.: Modeling the shape of the scene: A holistic
  representation of the spatial envelope. IJCV  (2001)

\bibitem{Palmer1981}
Palmer, S., Rosch, E., Chase, P.: Canonical perspective and the perception of
  objects. Attention and performance  (1981)

\bibitem{Raguram2009}
Raguram, R., Lazebnik, S.: Computing iconic summaries of general visual
  concepts. In: Internet Vision Workshop at CVPR (2009)

\bibitem{Raguram2011}
Raguram, R., Wu, C., Frahm, J.M., Lazebnik, S.: Modeling and recognition of
  landmark image collections using iconic scene graphs. IJCV  (2011)

\bibitem{Sanchez2013}
S{\'a}nchez, J., Perronnin, F., Mensink, T., Verbeek, J.: Image classification
  with the fisher vector: Theory and practice. IJCV  (2013)

\bibitem{Simon2007}
Simon, I., Snavely, N., Seitz, S.: Scene summarization for online image
  collections. In: ICCV (2007)

\bibitem{Wah2011}
Wah, C., Branson, S., Welinder, P., Perona, P., Belongie, S.: {The CUB-200-2011
  Dataset}. Tech. rep., CalTech (2011)

\bibitem{Wah2011iccv}
Wah, C., S.Branson, Perona, P., Belongie, S.: Multiclass recognition and part
  localization with humans in the loop. In: ICCV (2011)

\bibitem{Weinshall1994}
Weinshall, D., Werman, M., Gdalyahu, Y.: Canonical views, or the stability and
  likelihood of images of 3d objects. In: Image Understanding Workshop (1994)

\bibitem{Weyand2011}
Weyand, T., Leibe, B.: Discovering favorite views of popular places with
  iconoid shift. In: ICCV (2011)

\end{thebibliography}
